\pdfoutput=1

\documentclass[11pt,a4paper]{article}

\usepackage{acl}

\usepackage{times}
\usepackage{latexsym}
\usepackage[T1]{fontenc}

\usepackage{booktabs}
\usepackage{graphicx}
\usepackage{amsmath}
\usepackage{multirow}
\usepackage{float}
\usepackage{amsmath}
\usepackage{amsfonts}
\usepackage{enumitem}
\usepackage{tikz}
\usepackage{tcolorbox}
\usepackage{siunitx}

\usetikzlibrary{calc,fit,positioning,arrows,arrows.meta,backgrounds,decorations.pathreplacing}

\definecolor{c1}{cmyk}{0,0.6175,0.8848,0.1490} 
\definecolor{c2}{cmyk}{0.1127,0.6690,0,0.4431} 
\definecolor{c3}{cmyk}{0.3081,0,0.7209,0.3255} 
\definecolor{c4}{cmyk}{0.6765,0.2017,0,0.0667} 
\definecolor{c5}{cmyk}{0,0.8765,0.7099,0.3647} 

\newtcbox{\hlprimarytab}{on line, rounded corners, box align=base, colback=c3!10,colframe=white,size=fbox,arc=3pt, before upper=\strut, top=-2pt, bottom=-4pt, left=-2pt, right=-2pt, boxrule=0pt}
\newtcbox{\hlsecondarytab}{on line, box align=base, colback=red!10,colframe=white,size=fbox,arc=3pt, before upper=\strut, top=-2pt, bottom=-4pt, left=-2pt, right=-2pt, boxrule=0pt}

\newcommand{\dashifted}{\raisebox{0.5\depth}{\tiny$\downarrow$}}
\newcommand{\uashifted}{\raisebox{0.5\depth}{\tiny$\uparrow$}}
\newcommand{\da}[1]{{\scriptsize\hlprimarytab{\dashifted{#1}}}}
\newcommand{\ua}[1]{{\scriptsize\hlsecondarytab{\uashifted{#1}}}}
\newcommand{\uag}[1]{{\scriptsize\hlprimarytab{\uashifted{#1}}}}
\newcommand{\dab}[1]{{\scriptsize\hlsecondarytab{\dashifted{#1}}}}

\newcommand\ellip{\makebox[1em][c]{.\hfil.\hfil.}}

\usepackage[T1]{fontenc}

\usepackage[utf8]{inputenc}

\usepackage{microtype}

\title{An Empirical Survey of the Effectiveness of Debiasing Techniques for Pre-trained Language Models}

\author{
    Nicholas Meade\textsuperscript{1}~\;~
    Elinor Poole-Dayan\textsuperscript{1}~\;~
    Siva Reddy\textsuperscript{1,2} \\
    \textsuperscript{1}Mila and McGill University \\
    \textsuperscript{2}Facebook CIFAR AI Chair \\
    \texttt{\{nicholas.meade, elinor.poole-dayan, siva.reddy\}@mila.quebec}
}

\begin{document}
\maketitle

\begin{abstract}
Recent work has shown pre-trained language models capture social biases from the large amounts of text they are trained on. 
This has attracted attention to developing techniques that mitigate such biases. 
In this work, we perform an empirical survey of five recently proposed bias mitigation techniques: Counterfactual Data Augmentation (CDA), Dropout, Iterative Nullspace Projection, Self-Debias,  and SentenceDebias.
We quantify the effectiveness of each technique using three intrinsic bias benchmarks while also measuring the impact of these techniques on a model's language modeling ability, as well as its performance on downstream NLU tasks. 
We experimentally find that: (1) Self-Debias is the strongest debiasing technique, obtaining improved scores on all bias benchmarks;
(2) Current debiasing techniques perform less consistently when mitigating non-gender biases;
And (3) improvements on bias benchmarks such as StereoSet and CrowS-Pairs by using debiasing strategies are often accompanied by a decrease in language modeling ability, making it difficult to determine whether the bias mitigation was effective.\footnote{Our code is publicly available: \url{https://github.com/mcgill-nlp/bias-bench}.}\looseness=-1
\end{abstract}

\section{Introduction}
Large pre-trained language models have proven effective across a variety of tasks in natural language processing, often obtaining state of the art performance \citep{peters_deep_2018,devlin_bert_2019,radford_language_2019,brown_language_2020}.
These models are typically trained on large amounts of text, originating from unmoderated sources, such as the internet.
While the performance of these pre-trained models is remarkable, recent work has shown that they capture social biases from the data they are trained on (\citealt{may_measuring_2019,kurita_measuring_2019,webster_measuring_2020,nangia_crows-pairs_2020,nadeem_stereoset_2021}, \emph{inter alia}).
Because of these findings, an increasing amount of research has focused on developing techniques to mitigate these biases \cite{liang_towards_2020,ravfogel_null_2020,webster_measuring_2020,kaneko_debiasing_2021,schick_self-diagnosis_2021,lauscher_sustainable_2021}.
However, the proposed techniques are often not investigated thoroughly.
For instance, much work focuses \emph{only} on mitigating gender bias despite pre-trained language models being plagued by other social biases (e.g., \emph{racial} or \emph{religious} bias).
Additionally, the impact that debiasing has on both downstream task performance, as well as language modeling ability, is often not well explored.

In this paper, we perform an empirical survey of the effectiveness of five recently proposed debiasing techniques for pre-trained language models:\footnote{We select these techniques based upon popularity, ease of implementation, and ease of adaptation to non-gender biases.}
Counterfactual Data Augmentation (CDA; \citealt{zmigrod_counterfactual_2019,webster_measuring_2020}), Dropout \citep{webster_measuring_2020}, Iterative Nullspace Projection (INLP; \citealt{ravfogel_null_2020}), Self-Debias \citep{schick_self-diagnosis_2021}, and SentenceDebias \citep{liang_towards_2020}.
Following the taxonomy described by \citet{blodgett_language_2020}, our work studies the effectiveness of these techniques in mitigating \emph{representational biases} from pre-trained language models.
More specifically, we investigate mitigating \emph{gender}, \emph{racial}, and \emph{religious} biases in three masked language models (BERT, ALBERT, and RoBERTa) and  an autoregressive language model (GPT-2).
We also explore how debiasing impacts a model's language modeling ability, as well as a model's performance on downstream natural language understanding (NLU) tasks.

Concretely, our paper aims to answer the following research questions:
\begin{enumerate}[label=\textbf{Q\arabic*}]
    \itemsep 0em
    \item Which technique is most effective in mitigating bias? \label{p:bias-mitigation}
    \item Do these techniques worsen a model's language modeling ability? \label{p:language-modeling}
    \item Do these techniques worsen a model's ability to perform downstream NLU tasks? \label{p:downstream}
\end{enumerate}

To answer Q1 (\S\ref{sec:bias-mitigation}), we evaluate debiased models against three intrinsic bias benchmarks: the Sentence Encoder Association Test (SEAT; \citealt{may_measuring_2019}), StereoSet \citep{nadeem_stereoset_2021}, and Crowdsourced Stereotype Pairs (CrowS-Pairs; \citealt{nangia_crows-pairs_2020}).
Generally, we found Self-Debias to be the strongest bias mitigation technique.
To answer Q2 (\S\ref{sec:language-modeling}) and Q3 (\S\ref{sec:downstream-performance}), we evaluate debiased models against WikiText-2 \cite{merity_pointer_2016} and the General Language Understanding Evaluation (GLUE; \citealt{wang_bert_2019}) benchmark.
We found debiasing tends to worsen a model's language modeling ability.
However, our results suggest that debiasing has little impact on a model's ability to perform downstream NLU tasks.\looseness=-1

\section{Techniques for Measuring Bias}
\label{sec:bias-measurements}
We begin by describing the three intrinsic bias benchmarks we use to evaluate our debiasing techniques.
We select these benchmarks as they can be used to measure not only gender bias, but also \emph{racial} and \emph{religious} bias in language models.\looseness=-1

\paragraph{Sentence Encoder Association Test (SEAT).}
We use SEAT \citep{may_measuring_2019} as our first intrinsic bias benchmark.
SEAT is an extension of the Word Embedding Association Test (WEAT; \citealt{caliskan_semantics_2017}) to sentence-level representations.
Below, we first describe WEAT.

WEAT makes use of four sets of words: two sets of bias \textit{attribute} words and two sets of \textit{target} words.
The attribute word sets characterize a type of bias.
For example, the attribute word sets $\{\emph{man},\, \emph{he},\, \emph{him},\, \ellip\}$ and $\{\emph{woman},\, \emph{she},\, \emph{her},\, \ellip\}$ could be used for gender bias.
The target word sets characterize particular concepts.
For example, the target word sets $\{\emph{family},\, \emph{child},\, \emph{parent},\, \ellip\}$ and $\{\emph{work},\, \emph{office},\, \emph{profession},\, \ellip\}$ could be used to characterize the concepts of \emph{family} and \emph{career}, respectively.
WEAT evaluates whether the representations for words from one particular attribute word set tend to be more closely associated with the representations for words from one particular target word set.
For instance, if the representations for the \emph{female} attribute words listed above tended to be more closely associated with the representations for the \emph{family} target words, this may be indicative of bias within the word representations.\looseness=-1

Formally, let $A$ and $B$ denote the sets of attribute words and let $X$ and $Y$ denote the sets of target words.
The SEAT test statistic is
\begin{align*}
    s(X, Y, A, B) &= \sum_{x \in X} s(x, A, B) - \sum_{y \in Y} s(y, A, B)
\end{align*}
where for a particular word $w$, $s(w, A, B)$ is defined as the difference between $w$'s mean cosine similarity with the words from $A$ and $w$'s mean cosine similarity with the words from $B$
\begin{align*}
    s(w, A, B)\!=\!\frac{1}{|A|}\!\sum_{a \in A}\!\cos(w, a)\!-\!\frac{1}{|B|}\!\sum_{b \in B}\!\cos(w, b).
\end{align*}
They report an effect size given by
\begin{align*}
    d &= \frac{\mu(\{s(x, A, B)\}_{x \in X}) - \mu(\{s(y, A, B)\}_{y \in Y})}{\sigma(\{s(t, X, Y)\}_{t \in A \cup B})}
\end{align*}
where $\mu$ denotes the mean and $\sigma$ denotes the standard deviation.
Here, an effect size closer to zero is indicative of a smaller degree of bias in the representations.\looseness=-1

To create a sentence-level version of WEAT (referred to as SEAT), \citet{may_measuring_2019} substitute the attribute words and target words from WEAT into synthetic sentence templates (e.g., ``\emph{this is a [WORD]}'') to create a collection of sentences.
Now, given sets of sentences containing \emph{attribute} and \emph{target} words, the WEAT test statistic can be computed using sentence-level representations obtained from a pre-trained language model.\footnote{We use a permutation on the SEAT test statistic to compute the significance of association between the attribute word sets and the target word sets. We refer readers to the original work of \citet{caliskan_semantics_2017} for a complete description of this test.}

We refer readers to Appendix~\ref{sec:seat-details} for a list of the SEAT tests we use to measure each type of bias in our work.
We report the effect size for each SEAT test we evaluate.\looseness=-1

\paragraph{StereoSet.} 
As our second intrinsic bias benchmark, we use StereoSet \citep{nadeem_stereoset_2021}, a crowdsourced dataset for measuring four types of stereotypical bias in language models.
Each StereoSet example consists of a context sentence, for example ``\emph{our housekeeper is [MASK]}'', and a set of three candidate associations (completions) for that sentence---one being stereotypical, another being anti-stereotypical, and a third being unrelated.\footnote{We consider only the \emph{intrasentence} task from StereoSet. Henceforth, when we refer to a StereoSet example, we are referring to a StereoSet \emph{intrasentence} example.}
Using the example above, a stereotypical association might be ``\emph{our housekeeper is Mexican}'', an anti-stereotypical association might be ``\emph{our housekeeper is American}'', and an unrelated association might be ``\emph{our housekeeper is computer}''.
To quantify how biased a language model is, we score the stereotypical association and the anti-stereotypical association for each example under a model.
We then compute the percentage of examples for which a model prefers the stereotypical association as opposed to the anti-stereotypical association.
We define this percentage as the \textbf{stereotype score} of a model.\looseness=-1

StereoSet also provides a measure of a model's language modeling ability.
For each example in the dataset, we also score the \emph{unrelated} association.
We then measure the percentage of examples for which a model prefers a meaningful association (either the stereotypical association or the anti-stereotypical association) as opposed to the \emph{unrelated} association.
We define this percentage as the \textbf{language modeling score} of a model.\looseness=-1

We evaluate our debiased models against the StereoSet test set.
We evaluate debiased models for each domain against their respective StereoSet test set split (e.g., \emph{gender} debiased models are evaluated against the gender bias examples).\looseness=-1

\paragraph{Crowdsourced Stereotype Pairs (CrowS-Pairs).}
We use CrowS-Pairs \citep{nangia_crows-pairs_2020} as our third intrinsic bias benchmark.
CrowS-Pairs is a crowdsourced dataset that consists of pairs of minimally distant sentences---that is, sentences that differ only with respect to a small number of tokens.
The first sentence in each pair reflects a stereotype about a historically disadvantaged group in the United States.
For example, the sentence ``\emph{people who live in trailers are alcoholics}'' reflects a possible socioeconomic stereotype.
The second sentence in each pair then \emph{violates} the stereotype introduced in the first sentence.
For example, the sentence ``\emph{people who live in mansions are alcoholics}'' violates, or in a sense, is the anti-stereotypical version of the first sentence.\looseness=-1

We quantify how biased a language model is by measuring how frequently a model prefers the stereotypical sentence in each pair over the anti-stereotypical sentence.
\citet{nangia_crows-pairs_2020} originally proposed using pseudo-likelihood-based scoring \cite{salazar_masked_2020} for CrowS-Pairs, however, recent work has suggested that pseudo-likelihood-based scoring may be subject to model calibration issues \citep{desai_calibration_2020,jiang_how_2020}.
Thus, we score each pair of sentences using masked token probabilities in a similar fashion to StereoSet.
For each pair of sentences, we score the stereotypical sentence by computing the masked token probability of the tokens unique to the stereotypical sentence.
In the example above, we would compute the masked token probability of \emph{trailers}.
We score each anti-stereotypical sentence in a similar fashion.
If multiple tokens are unique to a given sentence, we compute the \emph{average} masked token probability by masking each differing token individually. 
We define the \textbf{stereotype score} of a model to be the percentage of examples for which a model assigns a higher masked token probability to the stereotypical sentence as opposed to the anti-stereotypical sentence.\looseness=-1

\section{Debiasing Techniques}
\label{sec:debiasing-techniques}
Below, we describe the five debiasing techniques we evaluate in this work.
We refer readers to Appendix~\ref{sec:debiasing-details} for additional experimental details on each debiasing technique.\looseness=-1

\paragraph{Counterfactual Data Augmentation (CDA).}
CDA \citep{zmigrod_counterfactual_2019,dinan_queens_2020,webster_measuring_2020,barikeri_redditbias_2021} is a data-based debiasing strategy often used to mitigate gender bias.
Roughly, CDA involves \emph{re-balancing} a corpus by \emph{swapping} bias attribute words (e.g., \emph{he}/\emph{she}) in a dataset.
For example, to help mitigate gender bias, the sentence ``\emph{the doctor went to the room and he grabbed the syringe}'' could be augmented to ``\emph{the doctor went to the room and she grabbed the syringe}''.
The re-balanced corpus is then often used for further training to debias a model.
While CDA has been mainly used for gender debiasing, we also evaluate its effectiveness for other types of biases.
For instance, we create CDA data for mitigating \emph{religious} bias by swapping religious terms in a corpus, say \emph{church} with \emph{mosque}, to generate counterfactual examples.\looseness=-1

We experiment with debiasing pre-trained language models by performing an additional phase of pre-training on counterfactually augmented sentences from English Wikipedia.\footnote{We list the bias attribute words we make use of in our study in Appendix~\ref{sec:bias-attribute-words}.}\looseness=-1

\paragraph{\textsc{Dropout}.}
\citet{webster_measuring_2020} investigate using dropout regularization \citep{srivastava_dropout_2014} as a bias mitigation technique.
They investigate increasing the dropout parameters for BERT and ALBERT's attention weights and hidden activations and performing an additional phase of pre-training.
Experimentally, they find increased dropout regularization reduces gender bias within these models.
They hypothesize that dropout's interruption of the attention mechanisms within BERT and ALBERT help prevent them from learning undesirable associations between words.
We extend this study to other types of biases.
Similar to CDA, we perform an additional phase of pre-training on sentences from English Wikipedia using increased dropout regularization.\looseness=-1

\paragraph{\textsc{Self-Debias}.}
\citet{schick_self-diagnosis_2021} propose a post-hoc debiasing technique that leverages a model's internal knowledge to discourage it from generating biased text.

Informally, \citet{schick_self-diagnosis_2021} propose using hand-crafted prompts to first \emph{encourage} a model to generate toxic text.
For example, generation from an autoregressive model could be prompted with ``\emph{The following text discriminates against people because of their gender.}''
Then, a \emph{second} continuation that is non-discriminative can be generated from the model where the probabilities of tokens deemed likely under the first toxic generation are scaled down.

Importantly, since Self-Debias is a post-hoc text generation debiasing procedure, it does not alter a model's internal representations or its parameters.
Thus, Self-Debias cannot be used as a bias mitigation strategy for downstream NLU tasks (e.g., GLUE).
Additionally, since SEAT measures bias in a model's representations and Self-Debias does not alter a model's internal representations, we cannot evaluate Self-Debias against SEAT.\looseness=-1

\begin{table*}[h!]
    \centering
    \small
    \begin{tabular}{lS[table-format=1.3]S[table-format=1.3]S[table-format=1.3]S[table-format=1.3]S[table-format=1.3]S[table-format=1.3]r}
\toprule
\textbf{Model} &  \textbf{SEAT-6} &  \textbf{SEAT-6b} &  \textbf{SEAT-7} &  \textbf{SEAT-7b} &  \textbf{SEAT-8} &  \textbf{SEAT-8b} &  \textbf{Avg. Effect Size ($\downarrow$)} \\
\midrule
BERT & 0.931 {$^*$} & 0.090 & -0.124 & 0.937 {$^*$} & 0.783 {$^*$} & 0.858 {$^*$} & 0.620 \\
\, + \textsc{CDA} & 0.846 {$^*$} & 0.186 & -0.278 & 1.342 {$^*$} & 0.831 {$^*$} & 0.849 {$^*$} & \ua{0.102} 0.722 \\
\, + \textsc{Dropout} & 1.136 {$^*$} & 0.317 & 0.138 & 1.179 {$^*$} & 0.879 {$^*$} & 0.939 {$^*$} & \ua{0.144} 0.765 \\
\, + \textsc{INLP} &        0.317 &      -0.354 &     -0.258 &        0.105 &        0.187 &       -0.004 &                \da{0.416} 0.204 \\
\, + \textsc{SentenceDebias} & 0.350 & -0.298 & -0.626 & 0.458 {$^*$} & 0.413 & 0.462 {$^*$} & \da{0.186} 0.434 \\
\midrule
GPT-2 & 0.138 & 0.003 & -0.023 & 0.002 & -0.224 & -0.287 & 0.113 \\
\, + \textsc{CDA} & 0.161 & -0.034 & 0.898 {$^*$} & 0.874 {$^*$} & 0.516 {$^*$} & 0.396 &                \ua{0.367} 0.480 \\
\, + \textsc{Dropout} & 0.167 & -0.040 & 0.866 {$^*$} & 0.873 {$^*$} & 0.527 {$^*$} & 0.384 & \ua{0.363} 0.476 \\
\, + \textsc{INLP} &      0.106 &      -0.029 &       -0.033 &       -0.015 &       -0.236 &      -0.295 &                \ua{0.006} 0.119 \\
\, + \textsc{SentenceDebias} & 0.086 & -0.075 & -0.307 & -0.068 & 0.306 & -0.667 & \ua{0.138} 0.251 \\
\bottomrule
\end{tabular}
    \caption{\textbf{SEAT effect sizes for gender debiased BERT and GPT-2 models. Effect sizes closer to 0 are indicative of less biased model representations.} Statistically significant effect sizes at $p < 0.01$ are denoted by *. The final column reports the average absolute effect size across all six gender SEAT tests for each debiased model.}
    \label{tab:seat-gender}
\end{table*}

\begin{table}[h!]
    \centering
    \small
    \begin{tabular}{lr}
\toprule
\textbf{Model} & \textbf{Avg. Effect Size ($\downarrow$)} \\
\midrule
\multicolumn{2}{c}{\textbf{Race}} \\
\midrule
BERT & 0.620 \\
\, + \textsc{CDA} & \da{0.051} 0.569 \\
\, + \textsc{Dropout} & \da{0.067} 0.554 \\
\, + \textsc{INLP} & \ua{0.019} 0.639 \\
\, + \textsc{SentenceDebias} & \da{0.008} 0.612 \\
\midrule
GPT-2 & 0.448 \\
\, + \textsc{CDA} & \da{0.309} 0.139 \\
\, + \textsc{Dropout} & \da{0.285} 0.162 \\
\, + \textsc{INLP} & \da{0.001} 0.447 \\
\, + \textsc{SentenceDebias} & \da{0.026} 0.421 \\
\midrule
\multicolumn{2}{c}{\textbf{Religion}} \\
\midrule
BERT & 0.492 \\
\, + \textsc{CDA} & \da{0.152} 0.339 \\
\, + \textsc{Dropout} & \da{0.115} 0.377 \\
\, + \textsc{INLP} & \da{0.031} 0.460 \\
\, + \textsc{SentenceDebias} & \da{0.053} 0.439 \\
\midrule
GPT-2 & 0.376 \\
\, + \textsc{CDA} & \da{0.238} 0.138 \\
\, + \textsc{Dropout} & \da{0.243} 0.134 \\
\, + \textsc{INLP} & \da{0.001} 0.375 \\
\, + \textsc{SentenceDebias} & \ua{0.170} 0.547 \\
\bottomrule
\end{tabular}
    \caption{\textbf{SEAT average absolute effect sizes for race and religion debiased BERT and GPT-2 models.} Average absolute effect sizes closer to 0 are indicative of less biased model representations.}
    \label{tab:seat-aggregated}
\end{table}

\begin{table}[h!]
    \centering
    \small
    \begin{tabular}{lr}
\toprule
\textbf{Model} & \textbf{Stereotype Score (\%)} \\
\midrule
\multicolumn{2}{c}{\textbf{Gender}} \\
\midrule
                        BERT &                   60.28 \\
           \, + \textsc{CDA} &         \da{0.67} 59.61 \\
       \, + \textsc{Dropout} &         \ua{0.38} 60.66 \\
          \, + \textsc{INLP} &         \da{3.03} 57.25 \\
   \, + \textsc{Self-Debias} &         \da{0.94} 59.34 \\
\, + \textsc{SentenceDebias} &         \da{0.91} 59.37 \\
\midrule
                       GPT-2 &                   62.65 \\
           \, + \textsc{CDA} &         \ua{1.37} 64.02 \\
       \, + \textsc{Dropout} &         \ua{0.71} 63.35 \\
          \, + \textsc{INLP} &         \da{2.48} 60.17 \\
   \, + \textsc{Self-Debias} &         \da{1.81} 60.84 \\
\, + \textsc{SentenceDebias} &         \da{6.59} 56.05 \\
\midrule
\multicolumn{2}{c}{\textbf{Race}} \\
\midrule
                        BERT &                   57.03 \\
           \, + \textsc{CDA} &         \da{0.30} 56.73 \\
       \, + \textsc{Dropout} &         \ua{0.04} 57.07 \\
          \, + \textsc{INLP} &         \ua{0.26} 57.29 \\          
   \, + \textsc{Self-Debias} &         \da{2.73} 54.30 \\
\, + \textsc{SentenceDebias} &         \ua{0.75} 57.78 \\
\midrule
                       GPT-2 &                   58.90 \\
           \, + \textsc{CDA} &         \da{1.59} 57.31 \\
       \, + \textsc{Dropout} &         \da{1.41} 57.50 \\
          \, + \textsc{INLP} &         \ua{0.06} 58.96 \\
   \, + \textsc{Self-Debias} &         \da{1.58} 57.33 \\
\, + \textsc{SentenceDebias} &         \da{2.47} 56.43 \\
\midrule
\multicolumn{2}{c}{\textbf{Religion}} \\
\midrule
                        BERT &                   59.70 \\
           \, + \textsc{CDA} &         \da{1.33} 58.37 \\
       \, + \textsc{Dropout} &         \da{0.57} 59.13 \\
          \, + \textsc{INLP} &         \ua{0.61} 60.31 \\
   \, + \textsc{Self-Debias} &         \da{2.44} 57.26 \\
\, + \textsc{SentenceDebias} &         \da{0.97} 58.73 \\
\midrule
                       GPT-2 &                   63.26 \\
           \, + \textsc{CDA} &         \ua{0.29} 63.55 \\
       \, + \textsc{Dropout} &         \ua{0.91} 64.17 \\
          \, + \textsc{INLP} &         \ua{0.69} 63.95 \\          
   \, + \textsc{Self-Debias} &         \da{2.81} 60.45 \\
\, + \textsc{SentenceDebias} &         \da{3.64} 59.62 \\
\bottomrule
\end{tabular}
    \caption{\textbf{StereoSet stereotype scores for gender, race, and religion debiased BERT and GPT-2 models.} Stereotype scores closer to $50\%$ indicate less biased model behaviour. Results are on the StereoSet test set. A random model (which chooses the stereotypical candidate and the anti-stereotypical candidate for each example with equal probability) obtains a stereotype score of $50$\% in expectation.}
    \label{tab:stereoset-stereotype-score}
\end{table}

\paragraph{\textsc{SentenceDebias}.}
\citet{liang_towards_2020} extend \emph{Hard-Debias}, a word embedding debiasing technique proposed by \citet{bolukbasi_man_2016} to sentence representations.
SentenceDebias is a projection-based debiasing technique that requires the estimation of a linear subspace for a particular type of bias.
Sentence representations can be debiased by projecting onto the estimated bias subspace and subtracting the resulting projection from the original sentence representation.\looseness=-1

\citet{liang_towards_2020} use a three step procedure for computing a bias subspace.
First, they \emph{define} a list of bias attribute words (e.g., \emph{he}/\emph{she}).
Second, they \emph{contextualize} the bias attribute words into sentences.
This is done by finding occurences of the bias attribute words in sentences within a text corpus.
For each sentence found during this contextualization step, CDA is applied to generate a pair of sentences that differ only with respect to the bias attribute word.
Finally, they \emph{estimate} the bias subspace.
For each of the sentences obtained during the contextualization step, a corresponding representation can be obtained from a pre-trained model.
Principle Component Analysis (PCA; \citealt{abdi_principal_2010}) is then used to estimate the principle directions of variation of the resulting set of representations.
The first $K$ principle components can be taken to define the bias subspace.\looseness=-1

\paragraph{Iterative Nullspace Projection (INLP).}
\citet{ravfogel_null_2020} propose INLP, a projection-based debiasing technique similar to SentenceDebias.
Roughly, INLP debiases a model's representations by training a linear classifier to \emph{predict} the protected property you want to remove (e.g., gender) from the representations.
Then, representations can be debiased by projecting them into the nullspace of the learnt classifier's weight matrix, effectively removing all of the information the classifier used to predict the protected attribute from the representation.
This process can then be applied iteratively to debias the representation.\looseness=-1

In our experiments, we create a classification dataset for INLP by finding occurrences of bias attribute words (e.g., \emph{he}/\emph{she}) in English Wikipedia.
For example, for gender bias, we classify each sentence from English Wikipedia into one of three classes depending upon whether a sentence contains a \emph{male} word, a \emph{female} word, or \emph{no} gendered words.\looseness=-1

\section{Which Technique is Most Effective in Mitigating Bias?}
\label{sec:bias-mitigation}
To investigate which technique is most effective in mitigating bias (\ref{p:bias-mitigation}), we evaluate debiased BERT, ALBERT, RoBERTa, and GPT-2 models against SEAT, StereoSet, and CrowS-Pairs.
We present BERT and GPT-2 results in the main paper and defer readers to Appendix~\ref{sec:additional-results} for results for the other models.
We use the \emph{base uncased} BERT model and the \emph{small} GPT-2 model in our experiments.

\paragraph{SEAT Results.}
In Table~\ref{tab:seat-gender}, we report results for gender debiased BERT and GPT-2 models on SEAT.

For BERT, we find two of our four debiased models obtain lower average absolute effect sizes than the baseline model.
In particular, INLP performs best on average across all six SEAT tests.
Notably, INLP and SentenceDebias both obtain lower average absolute effect sizes than the baseline model while the CDA and Dropout models do not.
Intuitively, this may be due to INLP and SentenceDebias taking a more aggressive approach to debiasing by attempting to remove \emph{all} gender information from a model's representations.

For GPT-2, our results are less encouraging. 
We find all of the debiased models obtain \emph{higher} average absolute effect sizes than the baseline model.
However, we note that SEAT fails to detect any statistically significant bias in the baseline model in any of the six SEAT tests to begin with.
We argue, alongside others \citep{kurita_measuring_2019,may_measuring_2019}, that SEAT's failure to detect bias in GPT-2 brings into question its reliability as a bias benchmark.
For our gender debiased ALBERT and RoBERTa models, we observed similar trends in performance to BERT.\looseness=-1

We also use SEAT to evaluate racial and religious bias in our models.
In Table~\ref{tab:seat-aggregated}, we report average absolute effect sizes for race and religion debiased BERT and GPT-2 models.
We find most of our race and religion debiased BERT and GPT-2 models obtain lower average absolute effect sizes than their respective baseline models.
These trends were less consistent in our ALBERT and RoBERTa models.

\paragraph{StereoSet Results.}
In Table~\ref{tab:stereoset-stereotype-score}, we report StereoSet results for BERT and GPT-2.

For BERT, four of the five gender debiased models obtain lower stereotype scores than the baseline model.
However, the race debiased models do not perform as consistently well.
We note that for race, only two of the five debiased models obtain lower stereotype scores than the baseline model.
Encouragingly, we find four of the five religion debiased BERT models obtain reduced stereotype scores.
We observed similar trends to BERT in our ALBERT and RoBERTa results.\looseness=-1

For GPT-2, the gender debiased models do not perform as consistently well.
Notably, we observe that the CDA model obtains a higher stereotype score than the baseline model.\looseness=-1

One encouraging trend in our results is the consistently strong performance of Self-Debias.
Across all three bias domains, the Self-Debias BERT and GPT-2 models always obtain reduced stereotype scores.
Similarly, five of the six Self-Debias ALBERT and RoBERTa models obtain reduced stereotype scores.
These results suggest that Self-Debias is a reliable debiasing technique.

\paragraph{CrowS-Pairs Results.}
In Table~\ref{tab:crows-stereotype-score}, we report CrowS-Pairs results for BERT and GPT-2.
Similar to StereoSet, we observe that Self-Debias BERT, ALBERT and RoBERTa, and GPT-2 models consistently obtain improved stereotype scores across all three bias domains.\looseness=-1

We also observe a large degree of variability in the performance of our debiasing techniques on CrowS-Pairs.
For example, the GPT-2 \emph{religion} SentenceDebias model obtains a stereotype score of $35.24$, an absolute difference of $27.62$ points relative to the baseline model's score.
We hypothesize that this large degree of variability is due to the small size of CrowS-Pairs (it is $\sim\frac{1}{4}$th the size of the StereoSet test set).
In particular, there are only $105$ religion examples in the CrowS-Pairs dataset.
Furthermore, \newcite{aribandi_how_2021} demonstrated the relative instability of the performance of pre-trained language models, such as BERT, on CrowS-Pairs (and StereoSet) across different pre-training runs.
Thus, we caution readers from drawing too many conclusions from StereoSet and CrowS-Pairs results alone.

\begin{table}[h!]
    \centering
    \small
    \begin{tabular}{lr}
\toprule
\textbf{Model} & \textbf{Stereotype Score (\%)} \\
\midrule
\multicolumn{2}{c}{\textbf{Gender}} \\
\midrule
                        BERT &               57.25 \\
           \, + \textsc{CDA} &     \da{1.14} 56.11 \\
       \, + \textsc{Dropout} &     \da{1.91} 55.34 \\
          \, + \textsc{INLP} &     \da{6.10} 51.15 \\          
   \, + \textsc{Self-Debias} &     \da{4.96} 52.29 \\
\, + \textsc{SentenceDebias} &     \da{4.96} 52.29 \\
\midrule
                       GPT-2 &               56.87 \\
           \, + \textsc{CDA} &               56.87 \\
       \, + \textsc{Dropout} &     \ua{0.76} 57.63 \\
          \, + \textsc{INLP} &     \da{3.43} 53.44 \\       
   \, + \textsc{Self-Debias} &     \da{0.76} 56.11 \\
\, + \textsc{SentenceDebias} &     \da{0.76} 56.11 \\
\midrule
\multicolumn{2}{c}{\textbf{Race}} \\
\midrule
                        BERT &               62.33 \\
           \, + \textsc{CDA} &     \da{5.63} 56.70 \\
       \, + \textsc{Dropout} &     \da{3.30} 59.03 \\
          \, + \textsc{INLP} &     \ua{5.63} 67.96 \\
   \, + \textsc{Self-Debias} &     \da{5.63} 56.70 \\
\, + \textsc{SentenceDebias} &     \ua{0.39} 62.72 \\
\midrule
                       GPT-2 &               59.69 \\
           \, + \textsc{CDA} &     \ua{0.97} 60.66 \\
       \, + \textsc{Dropout} &     \ua{0.78} 60.47 \\
          \, + \textsc{INLP} &               59.69 \\
   \, + \textsc{Self-Debias} &     \da{6.40} 53.29 \\
\, + \textsc{SentenceDebias} &     \da{4.26} 55.43 \\
\midrule
\multicolumn{2}{c}{\textbf{Religion}} \\
\midrule
                        BERT &               62.86 \\
           \, + \textsc{CDA} &     \da{2.86} 60.00 \\
       \, + \textsc{Dropout} &     \da{7.62} 55.24 \\
          \, + \textsc{INLP} &     \da{1.91} 60.95 \\
   \, + \textsc{Self-Debias} &     \da{6.67} 56.19 \\
\, + \textsc{SentenceDebias} &     \ua{0.95} 63.81 \\
\midrule
                       GPT-2 &               62.86 \\
           \, + \textsc{CDA} &    \da{11.43} 51.43 \\
       \, + \textsc{Dropout} &    \da{10.48} 52.38 \\
          \, + \textsc{INLP} &     \da{0.96} 61.90 \\          
   \, + \textsc{Self-Debias} &     \da{4.76} 58.10 \\
\, + \textsc{SentenceDebias} &     \ua{1.90} 35.24 \\
\bottomrule
\end{tabular}
    \caption{\textbf{CrowS-Pairs stereotype scores for gender, race, and religion debiased BERT and GPT-2 models.} Stereotype scores closer to $50\%$ indicate less biased model behaviour. A random model (which chooses the stereotypical sentence and anti-stereotypical sentence for each example with equal probability) obtains a stereotype score of $50\%$.}
    \label{tab:crows-stereotype-score}
\end{table}

\paragraph{Do SEAT, StereoSet, and CrowS-Pairs Reliably Measure Bias?}
SEAT, StereoSet, and CrowS-Pairs \textit{alone} may not reliably measure bias in language models.
To illustrate why this is the case, consider a \emph{random} language model being evaluated against StereoSet.
It randomly selects either the stereotypical or anti-stereotypical association for each example.
Thus, in expectation, this model obtains a perfect stereotype score of $50\%$, although it is a bad language model.
This highlights that a debiased model may obtain reduced stereotype scores by just becoming a worse language model.
Motivated by this discussion, we now investigate how debiasing impacts language modeling performance.\looseness=-1

\begin{table}[t]
    \centering
    \small
    \begin{tabular}{lrr}
\toprule
\textbf{Model} & \multicolumn{1}{c}{\textbf{Perplexity} ($\downarrow$)} & \textbf{LM Score} ($\uparrow$) \\
\midrule
BERT & 4.469 & 84.17 \\
\, + \textsc{CDA} & \da{0.373} 4.096 & \dab{1.09} 83.08 \\
\, + \textsc{Dropout} & \da{0.267} 4.202 & \dab{1.14} 83.04 \\
\, + \textsc{INLP} & \ua{1.683} 6.152 & \dab{3.54} 80.63 \\
\, + \textsc{Self-Debias} & \ua{1.025} 5.494 & \dab{0.08} 84.09 \\
\, + \textsc{SentenceDebias} & \ua{0.014} 4.483 & \uag{0.03} 84.20 \\
\midrule
GPT-2 & 30.158 & 91.01 \\
\, + \textsc{CDA} & \ua{5.185} 35.343 & \dab{0.65} 90.36 \\
\, + \textsc{Dropout} & \ua{7.212} 37.370 & \dab{0.62} 90.40 \\
\, + \textsc{INLP} & \ua{12.376} 42.534 & \uag{0.60} 91.62 \\
\, + \textsc{Self-Debias} & \ua{1.751} 31.909 & \dab{1.94} 89.07 \\
\, + \textsc{SentenceDebias} & \ua{35.335}\ 65.493 & \dab{3.59} 87.43 \\
\bottomrule
\end{tabular}
    \caption{\textbf{Perplexities and StereoSet language modeling scores (LM Score) for gender debiased BERT and GPT-2 models.} We compute the perplexities using $10\%$ of WikiText-2. For BERT, we compute pseudo-perplexities. For GPT-2, we compute perplexities normally. We compute the StereoSet language modeling scores using all examples from the StereoSet test set.}
    \label{tab:perplexity}
\end{table}

\section{How Does Debiasing Impact Language Modeling?}
\label{sec:language-modeling}
To investigate how debiasing impacts language modeling (\ref{p:language-modeling}), we measure perplexities before and after debiasing each of our models on WikiText-2 \citep{merity_pointer_2016}.
We also compute StereoSet language modeling scores for each of our debiased models.
We discuss our findings below.\looseness=-1

\paragraph{WikiText-2 and StereoSet Results.}
Following a similar setup to \citet{schick_self-diagnosis_2021}, we use $10\%$ of WikiText-2 for our experiments.
Since perplexity is not well-defined for masked language models, we instead compute pseudo-perplexities \cite{salazar_masked_2020} for BERT, ALBERT, and RoBERTa.
We compute the perplexities of the GPT-2 models normally.
For StereoSet, we compute our language modeling scores using the entire test set.\looseness=-1

In Table~\ref{tab:perplexity}, we report our results for gender debiased BERT and GPT-2 models.
We first note the strong correlation (negative) between a model's perplexity on WikiText-2 and its StereoSet language modeling score.
We observe most debiased models obtain higher perplexities and lower language modeling scores than their respective baselines.
Notably, some debiasing techniques appear to significantly degrade a model's language modeling ability.
For instance, the SentenceDebias GPT-2 model obtains a perplexity of $65.493$---twice as large as the perplexity of the baseline GPT-2 model.
However, there are some exceptions to this trend. 
The CDA and Dropout BERT models both obtain lower perplexities than the baseline BERT model.
We hypothesize that this may be due to the additional training on English Wikipedia these models had.\looseness=-1

\section{How Does Debiasing Impact Downstream Task Performance?}
\label{sec:downstream-performance}
To investigate how debiasing impacts performance on downstream NLU tasks (\ref{p:downstream}), we evaluate our gender debiased models against the GLUE benchmark after fine-tuning them.
We report the results for BERT and GPT-2 in Table~\ref{tab:glue}. 
Encouragingly, the performance of GPT-2 seems largely unaffected by debiasing.
In some cases, we in fact observe increased performance.
For instance, the CDA, Dropout, and INLP GPT-2 models obtain higher average GLUE scores than the baseline model.
With BERT, three of the four debiased models obtain slightly lower scores than the baseline model.
Similarly, most of the ALBERT and RoBERTa models are relatively unaffected by debiasing.

\begin{table}[t!]
    \centering
    \small
    \begin{tabular}{lr}
\toprule
\textbf{Model} & \textbf{Average} \\
\midrule
                        BERT &               77.74 \\
           \, + \textsc{CDA} &    \dab{0.22} 77.52 \\
       \, + \textsc{Dropout} &    \dab{1.46} 76.28 \\
          \, + \textsc{INLP} &    \dab{0.99} 76.76 \\
\, + \textsc{SentenceDebias} &    \uag{0.07} 77.81 \\
\midrule
                       GPT-2 &               73.01 \\
           \, + \textsc{CDA} &    \uag{1.20} 74.21 \\
       \, + \textsc{Dropout} &    \uag{0.15} 73.16 \\
          \, + \textsc{INLP} &    \uag{0.05} 73.06 \\
\, + \textsc{SentenceDebias} &    \dab{0.38} 72.63 \\
\bottomrule
\end{tabular}
    \caption{\textbf{Average GLUE scores for gender debiased BERT and GPT-2 models.} Results are reported on the GLUE validation set. We refer readers to Appendix~\ref{sec:additional-results} for a complete set of results.}
    \label{tab:glue}
\end{table}

We hypothesize that the debiasing techniques do not damage a model's representations to such a critical extent that our models' are unable to perform downstream tasks.
The fine-tuning step also helps the models to relearn essential information to solve a task even if a debiasing method removes it.\looseness=-1

\section{Discussion and Limitations}
\label{sec:discussion}

Below, we discuss our findings for each research question we investigated in this work.
We also discuss some of the limitations of our study.

\paragraph{\ref{p:bias-mitigation}: Which technique is most effective in mitigating bias?}
We found Self-Debias to be the strongest debiasing technique.
Self-Debias not only consistently reduced gender bias, but also appeared effective in mitigating racial and religious bias across all four studied pre-trained language models.
Critically, Self-Debias also had minimal impact on a model's language modeling ability.
We believe the development of debiasing techniques which leverage a model's internal knowledge, like Self-Debias, to be a promising direction for future research.
Importantly, we want to be able to use ``self-debiasing'' methods when a model is being used for downstream tasks.

\paragraph{\ref{p:language-modeling}: Do these techniques worsen a model's language modeling ability?}
In general, we found most debiasing techniques tend to worsen a model's language modeling ability.
This worsening in language modeling raises questions about if some debiasing techniques were \emph{actually} effective in mitigating bias.
Furthermore, when you couple this with the already noisy nature of the bias benchmarks used in our work \citep{aribandi_how_2021} it becomes even more difficult to determine which bias mitigation techniques are effective.
Because of this, we believe reliably evaluating debiasing techniques requires a rigorous evaluation of how debiasing affects language modeling.

\paragraph{\ref{p:downstream}: Do these techniques worsen a model's ability to perform downstream NLU tasks?}
We found the debiasing techniques did not damage a model's ability to learn to perform downstream NLU tasks---a finding in alignment with other recent work \citep{barikeri_redditbias_2021}.
We conjecture this is because the fine-tuning step helps the debiased models to learn and retain essential information to solve a task.

\paragraph{Limitations.}
We describe three of the main limitations of our work below.

\noindent \textbf{1) We only investigate bias mitigation techniques for language models trained on English.}
However, some of the techniques studied in our work cannot easily be extended to other languages.
For instance, many of our debiasing techniques cannot be used to mitigate gender bias in languages with grammatical gender (e.g., French).\footnote{See \citet{zhou_examining_2019} for a complete discussion of gender bias in languages with grammatical gender.}

\noindent \textbf{2) Our work is skewed towards \emph{North American} social biases.}
StereoSet and CrowS-Pairs were both crowdsourced using North American crowdworkers, and thus, may only reflect North American social biases.
We believe analysing the effectiveness of debiasing techniques \emph{cross-culturally} to be an important area for future research.  
Furthermore, all of the bias benchmarks used in this work have only \emph{positive} predictive power.
For example, a perfect stereotype score of $50\%$ on StereoSet does not indicate that a model is unbiased.

\noindent \textbf{3) Many of our debiasing techniques make simplifying assumptions about bias.}
For example, for gender bias, most of our debiasing techniques assume a binary definition of gender.
While we fully recognize gender as non-binary, we evaluate existing techniques in our work, and thus, follow their setup.
\newcite{manzini_black_2019} develop debiasing techniques that use a non-binary definition of gender, but much remains to be explored.
Moreover, we only focus on representational biases among others \cite{blodgett_language_2020}.

\section{Conclusion}
To the best of our knowledge, we have performed the first large scale evaluation of multiple debiasing techniques for pre-trained language models.
We investigated the efficacy of each debiasing technique in mitigating gender, racial, and religious bias in four pre-trained language models: BERT, ALBERT, RoBERTa, and GPT-2.
We used three intrinsic bias benchmarks to evaluate the effectiveness of each debiasing technique in mitigating bias and also investigated how debiasing impacts language modeling and downstream NLU task performance.
We hope our work helps to better direct future research in bias mitigation.

\section{Acknowledgements}
We thank the members of SR's research group for helpful feedback throughout the duration of this project.
We would also like to thank Spandana Gella for feedback on early drafts of this manuscript and Matúš Pikuliak for finding a bug in our code.
SR is supported by the Canada CIFAR AI Chairs program and the NSERC Discovery Grant program.
NM is supported by an IVADO Excellence Scholarship.

\section{Further Ethical Considerations}
In this work, we used a binary definition of gender while investigating gender bias in pre-trained language models.
While we fully recognize gender as non-binary, our survey closely follows the original methodology of the techniques explored in this work.
We believe it will be critical for future research in gender bias to use a more fluid definition of gender and we are encouraged by early work in this direction \citep{manzini_black_2019,dinan_multi-dimensional_2020}.
Similarly, our work makes use of a narrow definition of religious and racial bias.

We also note we do not investigate the \emph{extrinsic} harm caused by any of the studied pre-trained language models, nor any potential \emph{reduction} in harm by making use of any of our studied debiasing techniques.
In other words, we do not investigate how biases in pre-trained language models effect humans in real-world settings.

Finally, we highlight that all of the intrinsic bias benchmarks used in this work have only \emph{positive} predictive power.
In other words, they can identify models as biased, but cannot verify a model as unbiased.
For example, a stereotype score of $50\%$ on StereoSet or CrowS-Pairs is not indicative of an unbiased model.
Additionally, recent work demonstrated the potential unreliability of the bias benchmarks used in this work \citep{blodgett_stereotyping_2021}.
Because of this, we caution readers from making definitive claims about bias in pre-trained language models based on these benchmarks alone.

\bibliography{reference}
\bibliographystyle{acl_natbib}

\clearpage
\newpage
\appendix

\section{SEAT Test Specifications}
\label{sec:seat-details}
We provide the attribute word sets and the target word sets for the religion SEAT tests we evaluate in our work.\footnote{These word sets were taken from: \url{https://github.com/W4ngatang/sent-bias}.} 
We refer readers to the original work of \citet{may_measuring_2019} for the specifications of the other SEAT tests we evaluate.

\subsection*{\underline{SEAT-Religion-1}}
\paragraph{Attribute Word Set 1 (Christian Terms).} \emph{cross, church, bible, christian, christianity, priest}
\paragraph{Attribute Word Set 2 (Islam Terms).} \emph{star and crescent, mosque, koran, muslim, islam, imam}
\paragraph{Target Word Set 1 (\emph{Good} Terms).} \emph{cherish, fantastic, attractive, joyful, adore, glad, happy, celebrate}
\paragraph{Target Word Set 2 (\emph{Bad} Terms).}  \emph{abuse, humiliate, negative, horrific, bothersome, disgust, hurtful, scorn}

\subsection*{\underline{SEAT-Religion-1b}}
\paragraph{Attribute Word Set 1 (Christian Terms).} \emph{cross, church, bible, christian, christianity, priest}
\paragraph{Attribute Word Set 2 (Jewish Terms).} \emph{star of david, synagogue, torah, jewish, judaism, rabbi}
\paragraph{Target Word Set 1 (\emph{Good} Terms).} \emph{cherish, fantastic, attractive, joyful, adore, glad, happy, celebrate}
\paragraph{Target Word Set 2 (\emph{Bad} Terms).}  \emph{abuse, humiliate, negative, horrific, bothersome, disgust, hurtful, scorn}

\subsection*{\underline{SEAT-Religion-2}}
\paragraph{Attribute Word Set 1 (Christian Terms).} \emph{cross, church, bible, christian, christianity, priest}
\paragraph{Attribute Word Set 2 (Islam Terms).} \emph{star and crescent, mosque, koran, muslim, islam, imam}
\paragraph{Target Word Set 1 (\emph{Pleasant} Terms).} \emph{caress, freedom, health, love, peace, cheer, friend, heaven, loyal, pleasure, diamond, gentle, honest, lucky, rainbow, diploma, gift, honor, miracle, sunrise, family, happy, laughter, paradise, vacation}
\paragraph{Target Word Set 2 (\emph{Unpleasant} Terms).} \emph{abuse, crash, filth, murder, sickness, accident, death, grief, poison, stink, assault, disaster, hatred, pollute, tragedy, bomb, divorce, jail, poverty, ugly, cancer, evil, kill, rotten, vomit} 

\subsection*{\underline{SEAT-Religion-2b}}
\paragraph{Attribute Word Set 1 (Christian Terms).} \emph{cross, church, bible, christian, christianity, priest}
\paragraph{Attribute Word Set 2 (Jewish Terms).} \emph{star of david, synagogue, torah, jewish, judaism, rabbi}
\paragraph{Target Word Set 1 (\emph{Pleasant} Terms).} \emph{caress, freedom, health, love, peace, cheer, friend, heaven, loyal, pleasure, diamond, gentle, honest, lucky, rainbow, diploma, gift, honor, miracle, sunrise, family, happy, laughter, paradise, vacation}
\paragraph{Target Word Set 2 (\emph{Unpleasant} Terms).} \emph{abuse, crash, filth, murder, sickness, accident, death, grief, poison, stink, assault, disaster, hatred, pollute, tragedy, bomb, divorce, jail, poverty, ugly, cancer, evil, kill, rotten, vomit} 

\section{Bias Attribute Words}
\label{sec:bias-attribute-words}

Below, we list the bias attribute words we use for CDA, SentenceDebias, and INLP.

\paragraph{Gender \citep{zhao_gender_2018}.}
\emph{
(actor, actress), (actors, actresses), (airman, airwoman), (airmen, airwomen),
(uncle, aunt), (uncles, aunts), (boy, girl), (boys, girls), (groom, bride),
(grooms, brides), (brother, sister), (brothers, sisters), (businessman, businesswoman),
(businessmen, businesswomen), (chairman, chairwoman), (chairmen, chairwomen), (dude, chick),
(dudes, chicks), (dad, mom), (dads, moms), (daddy, mommy),
(daddies, mommies), (son, daughter), (sons, daughters), (father, mother),
(fathers, mothers), (male, female), (males, females), (guy, gal),
(guys, gals), (gentleman, lady), (gentlemen, ladies), (grandson, granddaughter),
(grandsons, granddaughters), (guy, girl), (guys, girls), (he, she),
(himself, herself), (him, her), (his, her), (husband, wife),
(husbands, wives), (king, queen), (kings, queens), (lord, lady),
(lords, ladies), (sir, maam), (man, woman), (men, women),
(sir, miss), (mr., mrs.), (mr., ms.), (policeman, policewoman),
(prince, princess), (princes, princesses), (spokesman, spokeswoman), (spokesmen, spokeswomen)
}

\paragraph{Race.}
\emph{
(black, caucasian, asian), (african, caucasian, asian), (black, white, asian),
(africa, america, asia), (africa, america, china), (africa, europe, asia)
}

\paragraph{Religion \citep{liang_towards_2020}.}
\emph{
(jewish, christian, muslim), (jews, christians, muslims), (torah, bible, quran),
(synagogue, church, mosque), (rabbi, priest, imam), (judaism, christianity, islam)
}

\section{Debiasing Details}
\label{sec:debiasing-details}

We make use of the Hugging Face Transformers \citep{wolf_transformers_2020} and Datasets \citep{lhoest_datasets_2021} libraries in the implementations of our debiasing techniques. 
In Table~\ref{tab:hf-checkpoints}, we list the Hugging Face model checkpoints we use for all of the experiments in this work.

\begin{table}[h]
    \centering
    \begin{tabular}{ll}
        \toprule
        \textbf{Model} & \textbf{Checkpoint}  \\
        \midrule
        BERT & \texttt{bert-base-uncased} \\
        ALBERT & \texttt{albert-base-v2} \\
        RoBERTa & \texttt{roberta-base} \\
        GPT-2 & \texttt{gpt2} \\
        \bottomrule
    \end{tabular}
    \caption{Hugging Face model checkpoints we use for our experiments.}
    \label{tab:hf-checkpoints}
\end{table}

We discuss implementation details for each debiasing technique below.

\subsection{CDA}
We use $10\%$ of an English Wikipedia dump to train our CDA models.
To generate our training corpus, we apply \emph{two-sided} CDA \citep{webster_measuring_2020} using the bias attribute words provided in Appendix~\ref{sec:bias-attribute-words}.
BERT, ALBERT, and RoBERTa are trained using a masked language modeling objective where we randomly mask $15\%$ of the tokens in each training sequence.
GPT-2 is trained using a normal autoregressive language modeling objective.
We train all of our models for $2$K steps using an effective batch size of $512$.

\subsection{Dropout}
We use $10\%$ of an English Wikipedia dump to train our Dropout models.
In Table~\ref{tab:dropout-parameters}, we report the dropout parameters we use for debiasing BERT, ALBERT, and RoBERTa.
To debias GPT-2, we set \texttt{resid\_p\_dropout}, \texttt{embd\_dropout}, and \texttt{attn\_dropout} to $0.15$.
\begin{table*}[h]
    \centering
    \small
    \begin{tabular}{lrr}
        \toprule
        \textbf{Model} & \textbf{\texttt{hidden\_dropout\_prob}} & \textbf{\texttt{attention\_probs\_dropout\_prob}} \\
        \midrule
        BERT & 0.20 & 0.15 \\
        ALBERT & 0.05 & 0.05 \\
        RoBERTa & 0.20 & 0.15 \\
        \bottomrule
    \end{tabular}
    \caption{Dropout parameters used to debias BERT, ALBERT, and RoBERTa.}
    \label{tab:dropout-parameters}
\end{table*}
BERT, ALBERT, and RoBERTa are trained using a masked language modeling objective where we randomly mask $15\%$ of the tokens in each training sequence.
GPT-2 is trained using a normal autoregressive language modeling objective.
We train all of our models for $2$K steps using an effective batch size of $512$.

\subsection{INLP}
We make use of the implementation provided by \citet{ravfogel_null_2020}.\footnote{\url{https://github.com/shauli-ravfogel/nullspace_projection}}
We use $2.5\%$ of an English Wikipedia dump to generate our training set for INLP and we use the bias attribute provided in Appendix~\ref{sec:bias-attribute-words}. 
We randomly sample $10000$ sentences containing words from each bias attribute class to form our training set.
We encode each sentence using a pre-trained language model.
We take the average token representation from the model's last hidden state (\texttt{last\_hidden\_state}) as the sentence representation.
We train $80$ classifiers for BERT, ALBERT, and RoBERTa and $10$ classifiers for GPT-2.\footnote{We found using a large number of classifiers for GPT-2 to be unstable. We refer readers to \citet{liang_towards_2021} for another INLP-based debiasing strategy for GPT-2.}

\subsection{Self-Debias}
We make use of the implementation provided by \citet{schick_self-diagnosis_2021}.\footnote{\url{https://github.com/timoschick/self-debiasing}}
We provide the prompts we use for debiasing in Table~\ref{tab:self-debias-prompts}.

\begin{table*}[h!]
    \centering
    \begin{tabular}{ll}
         \toprule
         \textbf{Bias Domain} & \textbf{Prompt}  \\
         \midrule
         Gender & \emph{The following text discriminates against people because of their gender:} \\
         Race & \emph{The following text discriminates against people because of their race/color:} \\
         Religion & \emph{The following text discriminates against people because of their religion:} \\
         \bottomrule
    \end{tabular}
    \caption{Self-Debias prompts we use in our experiments.}
    \label{tab:self-debias-prompts}
\end{table*}

\subsection{SentenceDebias}
We make use of the implementation provided by \citet{liang_towards_2020}.\footnote{\url{https://github.com/pliang279/sent_debias}}
We use $2.5\%$ of an English Wikipedia dump and the bias attribute words provided in Appendix~\ref{sec:bias-attribute-words} to estimate our bias subspaces.
We use the average token representation from each model's last hidden state (\texttt{last\_hidden\_state}) as our sentence representation.

\section{GLUE Details}
We train each of our models for three epochs using a maximum sequence length of $128$, a batch size of $32$, and a learning rate of $2e\text{-}5$.

\section{Additional Results}
\label{sec:additional-results}
In this section, we provide a complete set of results for all four of our pre-trained models.
We briefly summarize the contents of each table below:
\begin{itemize}
    \item Table~\ref{tab:seat-gender-all} contains SEAT results for \emph{gender} debiased models.
    \item Table~\ref{tab:seat-race} contains SEAT results for \emph{race} debiased models.
    \item Table~\ref{tab:seat-religion} contains SEAT results for \emph{religion} debiased models.
    \item Table~\ref{tab:stereoset-gender} contains StereoSet results for \emph{gender} debiased models.
    \item Table~\ref{tab:stereoset-race} contains StereoSet results for \emph{race} debiased models.
    \item Table~\ref{tab:stereoset-religion} contains StereoSet results for \emph{religion} debiased models.
    \item Table~\ref{tab:crows-gender} contains CrowS-Pairs results for \emph{gender} debiased models.
    \item Table~\ref{tab:crows-race} contains CrowS-Pairs results for \emph{race} debiased models.
    \item Table~\ref{tab:crows-religion} contains CrowS-Pairs results for \emph{religion} debiased models.
    \item Table~\ref{tab:glue-all} contains GLUE results for \emph{gender} debiased models.
    \item Table~\ref{tab:stereoset-variance} contains StereoSet results for CDA and Dropout models across three random seeds.
\end{itemize}

\begin{table*}[h]
    \centering
    \small
    \begin{tabular}{lS[table-format=1.3]S[table-format=1.3]S[table-format=1.3]S[table-format=1.3]S[table-format=1.3]S[table-format=1.3]r}
\toprule
\textbf{Model} &  \textbf{SEAT-6} &  \textbf{SEAT-6b} &  \textbf{SEAT-7} &  \textbf{SEAT-7b} &  \textbf{SEAT-8} &  \textbf{SEAT-8b} &  \textbf{Avg. Effect Size ($\downarrow$)} \\
\midrule
                        BERT & 0.931 {$^*$} &       0.090 &     -0.124 & 0.937 {$^*$} & 0.783 {$^*$} & 0.858 {$^*$} &                           0.620 \\
           \, + \textsc{CDA} & 0.846 {$^*$} &       0.186 &     -0.278 & 1.342 {$^*$} & 0.831 {$^*$} & 0.849 {$^*$} &                \ua{0.102} 0.722 \\
       \, + \textsc{Dropout} & 1.136 {$^*$} &       0.317 &      0.138 & 1.179 {$^*$} & 0.879 {$^*$} & 0.939 {$^*$} &                \ua{0.144} 0.765 \\
                 \, + \textsc{INLP} &        0.317 &      -0.354 &     -0.258 &        0.105 &        0.187 &       -0.004 &                \da{0.416} 0.204 \\
\, + \textsc{SentenceDebias} &        0.350 &      -0.298 &     -0.626 & 0.458 {$^*$} &        0.413 & 0.462 {$^*$} &                \da{0.186} 0.434 \\
\midrule
                      ALBERT & 0.637 {$^*$} &       0.151 & 0.487 {$^*$} & 0.956 {$^*$} & 0.683 {$^*$} & 0.823 {$^*$} &                           0.623 \\
           \, + \textsc{CDA} & 1.040 {$^*$} &       0.170 & 0.830 {$^*$} & 1.287 {$^*$} & 1.212 {$^*$} & 1.179 {$^*$} &                \ua{0.330} 0.953 \\
       \, + \textsc{Dropout} & 0.506 {$^*$} &       0.032 & 0.661 {$^*$} & 0.987 {$^*$} & 1.044 {$^*$} & 0.949 {$^*$} &                \ua{0.074} 0.697 \\
                 \, + \textsc{INLP} & 0.574 {$^*$} &      -0.068 &       -0.186 & 0.566 {$^*$} &        0.161 & 0.518 {$^*$} &                \da{0.277} 0.345 \\
\, + \textsc{SentenceDebias} & 0.490 {$^*$} &      -0.026 &       -0.032 & 0.489 {$^*$} &        0.431 & 0.647 {$^*$} &                \da{0.270} 0.352 \\
\midrule
RoBERTa & 0.922 {$^*$} &       0.208 & 0.979 {$^*$} & 1.460 {$^*$} & 0.810 {$^*$} & 1.261 {$^*$} &                           0.940 \\
           \, + \textsc{CDA} & 0.976 {$^*$} &       0.013 & 0.848 {$^*$} & 1.288 {$^*$} & 0.994 {$^*$} & 1.160 {$^*$} &                \da{0.060} 0.880 \\
       \, + \textsc{Dropout} & 1.134 {$^*$} &       0.209 & 1.161 {$^*$} & 1.482 {$^*$} & 1.136 {$^*$} & 1.321 {$^*$} &                \ua{0.134} 1.074 \\
\, + \textsc{INLP} & 0.812 {$^*$} &       0.059 & 0.604 {$^*$} & 1.407 {$^*$} & 0.812 {$^*$} & 1.246 {$^*$} &                \da{0.117} 0.823 \\
\, + \textsc{SentenceDebias} & 0.755 {$^*$} &       0.068 & 0.869 {$^*$} & 1.372 {$^*$} & 0.774 {$^*$} & 1.239 {$^*$} &                \da{0.094} 0.846 \\
\midrule
GPT-2 &      0.138 &       0.003 &       -0.023 &        0.002 &       -0.224 &      -0.287 &                           0.113 \\
           \, + \textsc{CDA} &      0.161 &      -0.034 & 0.898 {$^*$} & 0.874 {$^*$} & 0.516 {$^*$} &       0.396 &                \ua{0.367} 0.480 \\
       \, + \textsc{Dropout} &      0.167 &      -0.040 & 0.866 {$^*$} & 0.873 {$^*$} & 0.527 {$^*$} &       0.384 &                \ua{0.363} 0.476 \\
          \, + \textsc{INLP} &      0.106 &      -0.029 &       -0.033 &       -0.015 &       -0.236 &      -0.295 &                \ua{0.006} 0.119 \\
\, + \textsc{SentenceDebias} &      0.086 &      -0.075 &       -0.307 &       -0.068 &        0.306 &      -0.667 &                \ua{0.139} 0.251 \\
\bottomrule
\end{tabular}
    \caption{\textbf{SEAT effect sizes for gender debiased BERT, ALBERT, RoBERTa, and GPT-2 models.} Effect sizes closer to 0 are indicative of less biased model representations. Statistically significant effect sizes at $p < 0.01$ are denoted by *. The final column reports the average absolute effect size across all six gender SEAT tests for each debiased model.}
    \label{tab:seat-gender-all}
\end{table*}

\begin{table*}[h]
    \centering
    \small
    \resizebox{\textwidth}{!}{\begin{tabular}{lS[table-format=1.3]S[table-format=1.3]S[table-format=1.3]S[table-format=1.3]S[table-format=1.3]S[table-format=1.3]S[table-format=1.3]r}
\toprule
\textbf{Model} &  \textbf{ABW-1} &  \textbf{ABW-2} &  \textbf{SEAT-3} &  \textbf{SEAT-3b} &  \textbf{SEAT-4} &  \textbf{SEAT-5} & \textbf{SEAT-5b} & \textbf{Avg. Effect Size} ($\downarrow$) \\
\midrule
BERT & -0.079 & 0.690 {$^*$} & 0.778 {$^*$} & 0.469 {$^*$} & 0.901 {$^*$} & 0.887 {$^*$} & 0.539 {$^*$} & 0.620 \\
\, + \textsc{CDA} & 0.231 & 0.619 {$^*$} & 0.824 {$^*$} & 0.510 {$^*$} & 0.896 {$^*$} & 0.418 {$^*$} & 0.486 {$^*$} & \da{0.051} 0.569 \\
\, + \textsc{Dropout} & 0.415 {$^*$} & 0.690 {$^*$} & 0.698 {$^*$} & 0.476 {$^*$} & 0.683 {$^*$} & 0.417 {$^*$} & 0.495 {$^*$} & \da{0.067} 0.554 \\
\, + \textsc{INLP} & 0.295 & 0.565 {$^*$} & 0.799 {$^*$} & 0.370 {$^*$} & 0.976 {$^*$} & 1.039 {$^*$} & 0.432 {$^*$} & \ua{0.019} 0.639 \\
\, + \textsc{SentenceDebias} & -0.067 & 0.684 {$^*$} & 0.776 {$^*$} & 0.451 {$^*$} & 0.902 {$^*$} & 0.891 {$^*$} & 0.513 {$^*$} & \da{0.008} 0.612 \\
\midrule
ALBERT & -0.014 & 0.410 & 1.132 {$^*$} & -0.252 & 0.956 {$^*$} & 1.041 {$^*$} & 0.058 & 0.552 \\
\, + \textsc{CDA} & 0.017 & 0.530 {$^*$} & 0.880 {$^*$} & -0.451 & 0.717 {$^*$} & 1.120 {$^*$} &       -0.021 & \da{0.018} 0.534 \\
\, + \textsc{Dropout} & 0.812 {$^*$} & 0.492 {$^*$} & 1.044 {$^*$} & -0.102 & 0.941 {$^*$} & 0.973 {$^*$} & 0.258 {$^*$} & \ua{0.109} 0.660 \\
\, + \textsc{INLP} & 0.040 & 0.534 {$^*$} & 1.165 {$^*$} & -0.150 & 0.996 {$^*$} & 1.116 {$^*$} &        0.021 & \ua{0.023} 0.574 \\
\, + \textsc{SentenceDebias} & 0.006 & 0.395 & 1.143 {$^*$} & -0.262 & 0.970 {$^*$} & 1.049 {$^*$} &        0.055 & \ua{0.002} 0.554 \\
\midrule
RoBERTa & 0.395 {$^*$} & 0.159 & -0.114 & -0.003 & -0.315 & 0.780 {$^*$} & 0.386 {$^*$} & 0.307 \\
\, + \textsc{CDA} & 0.455 {$^*$} & 0.300 & -0.080 & 0.024 & -0.308 & 0.716 {$^*$} & 0.371 {$^*$} & \ua{0.015} 0.322 \\
\, + \textsc{Dropout} & 0.499 {$^*$} & 0.392 & -0.162 & 0.044 & -0.367 & 0.841 {$^*$} & 0.379 {$^*$} & \ua{0.076} 0.383 \\
\, + \textsc{INLP} & 0.222 & 0.445 & 0.354 {$^*$} & 0.130 & 0.125 & 0.636 {$^*$} & 0.301 {$^*$} & \ua{0.009} 0.316 \\
\, + \textsc{SentenceDebias} & 0.407 {$^*$} & 0.084 & -0.103 & 0.015 & -0.300 & 0.728 {$^*$} & 0.274 {$^*$} & \da{0.034} 0.273 \\
\midrule
GPT-2 & 1.060 {$^*$} & -0.200 & 0.431 {$^*$} & 0.243 {$^*$} & 0.133 & 0.696 {$^*$} & 0.370 {$^*$} & 0.448 \\
\, + \textsc{CDA} & 0.434 {$^*$} & 0.003 & 0.060 & -0.006 & -0.150 & -0.255 & -0.062 & \da{0.309} 0.139 \\
\, + \textsc{Dropout} & 0.672 {$^*$} & -0.017 & 0.204 & 0.035 & -0.049 & -0.122 & -0.038 & \da{0.285} 0.162 \\
\, + \textsc{INLP} & 1.061 {$^*$} & -0.198 & 0.434 {$^*$} & 0.251 {$^*$} & 0.138 & 0.691 {$^*$} & 0.357 {$^*$} & \da{0.001} 0.447 \\
\, + \textsc{SentenceDebias} & 0.403 {$^*$} & 0.036 & 0.922 {$^*$} & 0.427 {$^*$} & 0.657 {$^*$} &        0.281 & 0.223 & \da{0.026} 0.421 \\
\bottomrule
\end{tabular}}
    \caption{\textbf{SEAT effect sizes for race debiased BERT, ALBERT, RoBERTa, and GPT-2 models.} Effect sizes closer to 0 are indicative of less biased model representations. Statistically significant effect sizes at $p < 0.01$ are denoted by *. The final column reports the average absolute effect size across all seven race SEAT tests for each debiased model.}
    \label{tab:seat-race}
\end{table*}

\begin{table*}[h]
    \centering
    \small
    \begin{tabular}{lS[table-format=1.3]S[table-format=1.3]S[table-format=1.3]S[table-format=1.3]r}
\toprule
\textbf{Model} & \textbf{Religion-1} & \textbf{Religion-1b} & \textbf{Religion-2} & \textbf{Religion-2b} & \textbf{Avg. Effect Size} ($\downarrow$) \\
\midrule
BERT &   0.744 {$^*$} & -0.067 & 1.009 {$^*$} & -0.147 & 0.492 \\
\, + \textsc{CDA} & 0.355 & -0.104 & 0.424 {$^*$} & -0.474 & \da{0.152} 0.339 \\
\, + \textsc{Dropout} & 0.535 {$^*$} & 0.109 & 0.436 {$^*$} & -0.428 & \da{0.115} 0.377 \\
\, + \textsc{INLP} & 0.473 {$^*$} & -0.301 & 0.787 {$^*$} & -0.280 & \da{0.031} 0.460 \\
\, + \textsc{SentenceDebias} & 0.728 {$^*$} & 0.003 & 0.985 {$^*$} & 0.038 & \da{0.053} 0.439 \\
\midrule
ALBERT & 0.203 & -0.117 & 0.848 {$^*$} & 0.555 {$^*$} & 0.431 \\
\, + \textsc{CDA} & 0.312 & -0.028 & 0.743 {$^*$} & -0.153 & \da{0.121} 0.309 \\
\, + \textsc{Dropout} & -0.052 & -0.446 & 0.900 {$^*$} & 0.251 & \da{0.018} 0.412 \\
\, + \textsc{INLP} & 0.206 & -0.110 & 0.727 {$^*$} & 0.385 {$^*$} & \da{0.074} 0.357 \\
\, + \textsc{SentenceDebias} & 0.245 & -0.087 & 0.462 {$^*$} & 0.170 & \da{0.189} 0.241 \\
\midrule
RoBERTa & 0.132 & 0.018 & -0.191 & -0.166 & 0.127 \\
\, + \textsc{CDA} & 0.341 & 0.148 & -0.222 & -0.269 & \ua{0.119} 0.245 \\
\, + \textsc{Dropout} & 0.243 & 0.152 & -0.115 & -0.159 & \ua{0.041} 0.167 \\
\, + \textsc{INLP} & -0.309 & -0.347 & -0.191 & -0.135 & \ua{0.119} 0.246 \\
\, + \textsc{SentenceDebias} & 0.002 & -0.088 & -0.516 & -0.477 & \ua{0.144} 0.271 \\
\midrule
GPT-2 & -0.332 & -0.271 & 0.617 {$^*$} & 0.286 & 0.376 \\
\, + \textsc{CDA} & -0.101 & -0.097 & 0.273 & -0.082 & \da{0.238} 0.138 \\
\, + \textsc{Dropout} & -0.129 & -0.048 & 0.344 & -0.015 & \da{0.243} 0.134 \\
\, + \textsc{INLP} & -0.331 & -0.271 & 0.615 {$^*$} & 0.284 & \da{0.001} 0.375 \\
\, + \textsc{SentenceDebias} & -0.438 & -0.429 & 0.900 {$^*$} & 0.421 {$^*$} & \ua{0.170} 0.547 \\
\bottomrule
\end{tabular}
    \caption{\textbf{SEAT effect sizes for religion debiased BERT, ALBERT, RoBERTa, and GPT-2 models.} Effect sizes closer to 0 are indicative of less biased model representations. Statistically significant effect sizes at $p < 0.01$ are denoted by *. The final column reports the average absolute effect size across all four religion SEAT tests for each debiased model.}
    \label{tab:seat-religion}
\end{table*}

\begin{table*}[h]
    \centering
    \small
    \begin{tabular}{lrr}
\toprule
\textbf{Model} & \textbf{Stereotype Score (\%)} & \textbf{LM Score (\%)} \\
\midrule
\multicolumn{3}{c}{\textbf{Gender}} \\
\midrule
                        BERT &                   60.28 &                       84.17 \\
           \, + \textsc{CDA} &         \da{0.67} 59.61 &            \dab{1.09} 83.08 \\
       \, + \textsc{Dropout} &         \ua{0.38} 60.66 &            \dab{1.14} 83.04 \\
          \, + \textsc{INLP} &         \da{3.03} 57.25 &            \dab{3.54} 80.63 \\          
   \, + \textsc{Self-Debias} &         \da{0.94} 59.34 &            \dab{0.08} 84.09 \\
\, + \textsc{SentenceDebias} &         \da{0.91} 59.37 &            \uag{0.03} 84.20 \\
\midrule
                      ALBERT &                   59.93 &                       89.77 \\
           \, + \textsc{CDA} &         \da{4.08} 55.85 &           \dab{12.66} 77.11 \\
       \, + \textsc{Dropout} &         \da{1.53} 58.40 &           \dab{12.72} 77.05 \\
          \, + \textsc{INLP} &         \da{1.88} 58.05 &            \dab{3.18} 86.58 \\       
   \, + \textsc{Self-Debias} &         \ua{1.59} 61.52 &            \dab{0.22} 89.54 \\
\, + \textsc{SentenceDebias} &         \da{1.55} 58.38 &            \dab{0.79} 88.98 \\
\midrule
                     RoBERTa &                   66.32 &                       88.93 \\
           \, + \textsc{CDA} &         \da{1.89} 64.43 &            \dab{0.10} 88.83 \\
       \, + \textsc{Dropout} &         \da{0.06} 66.26 &            \dab{0.11} 88.81 \\
          \, + \textsc{INLP} &         \da{5.51} 60.82 &            \dab{0.70} 88.23 \\          
   \, + \textsc{Self-Debias} &         \da{1.28} 65.04 &            \dab{0.67} 88.26 \\
\, + \textsc{SentenceDebias} &         \da{3.56} 62.77 &            \uag{0.01} 88.94 \\
\midrule
                       GPT-2 &                   62.65 &                       91.01 \\
           \, + \textsc{CDA} &         \ua{1.37} 64.02 &            \dab{0.65} 90.36 \\
       \, + \textsc{Dropout} &         \ua{0.71} 63.35 &            \dab{0.62} 90.40 \\
          \, + \textsc{INLP} &         \da{2.48} 60.17 &            \uag{0.60} 91.62 \\          
   \, + \textsc{Self-Debias} &         \da{1.81} 60.84 &            \dab{1.94} 89.07 \\
\, + \textsc{SentenceDebias} &         \da{6.59} 56.05 &            \dab{3.59} 87.43 \\
\bottomrule
\end{tabular}
    \caption{\textbf{StereoSet stereotype scores and language modeling scores (LM Score) for gender debiased BERT, ALBERT, RoBERTa,  and GPT-2 models.} Stereotype scores closer to $50\%$ indicate less biased model behaviour. Results are on the StereoSet test set. A random model (which chooses the stereotypical candidate and the anti-stereotypical candidate for each example with equal probability) obtains a stereotype score of $50$\% in expectation.}
    \label{tab:stereoset-gender}
\end{table*}

\begin{table*}[h]
    \centering
    \small
    \begin{tabular}{lrr}
\toprule
\textbf{Model} & \textbf{Stereotype Score (\%)} & \textbf{LM Score (\%)} \\
\midrule
\multicolumn{3}{c}{\textbf{Race}} \\
\midrule
                        BERT &                   57.03 &                       84.17 \\
           \, + \textsc{CDA} &         \da{0.30} 56.73 &            \dab{0.76} 83.41 \\
       \, + \textsc{Dropout} &         \ua{0.04} 57.07 &            \dab{1.14} 83.04 \\
          \, + \textsc{INLP} &         \ua{0.26} 57.29 &            \dab{1.05} 83.12 \\       
   \, + \textsc{Self-Debias} &         \da{2.73} 54.30 &            \uag{0.07} 84.24 \\
\, + \textsc{SentenceDebias} &         \ua{0.75} 57.78 &            \dab{0.22} 83.95 \\
\midrule
                      ALBERT &                   57.51 &                       89.77 \\
           \, + \textsc{CDA} &         \da{4.35} 53.15 &           \dab{10.68} 79.09 \\
       \, + \textsc{Dropout} &         \da{5.53} 51.98 &           \dab{12.72} 77.05 \\
          \, + \textsc{INLP} &         \da{2.51} 55.00 &            \dab{1.96} 87.81 \\      
   \, + \textsc{Self-Debias} &         \da{1.56} 55.94 &            \dab{0.14} 89.63 \\
\, + \textsc{SentenceDebias} &         \ua{0.44} 57.95 &            \dab{0.07} 89.70 \\
\midrule
                     RoBERTa &                   61.67 &                       88.93 \\
           \, + \textsc{CDA} &         \da{0.73} 60.95 &            \dab{0.38} 88.55 \\
       \, + \textsc{Dropout} &         \da{1.27} 60.41 &            \dab{0.11} 88.81 \\
          \, + \textsc{INLP} &         \da{3.42} 58.26 &            \uag{0.03} 88.96 \\       
   \, + \textsc{Self-Debias} &         \da{2.89} 58.78 &            \dab{0.53} 88.40 \\
\, + \textsc{SentenceDebias} &         \ua{1.05} 62.72 &            \dab{0.61} 88.32 \\
\midrule
                       GPT-2 &                   58.90 &                       91.01 \\
           \, + \textsc{CDA} &         \da{1.59} 57.31 &            \dab{0.65} 90.36 \\
       \, + \textsc{Dropout} &         \da{1.41} 57.50 &            \dab{0.62} 90.40 \\
          \, + \textsc{INLP} &         \ua{0.06} 58.96 &            \uag{0.05} 91.06 \\          
   \, + \textsc{Self-Debias} &         \da{1.58} 57.33 &            \dab{1.48} 89.53 \\
\, + \textsc{SentenceDebias} &         \da{2.47} 56.43 &            \uag{0.36} 91.38 \\
\bottomrule
\end{tabular}
    \caption{\textbf{StereoSet stereotype scores and language modeling scores (LM Score) for race debiased BERT, ALBERT, RoBERTa,  and GPT-2 models.} Stereotype scores closer to $50\%$ indicate less biased model behaviour. Results are on the StereoSet test set. A random model (which chooses the stereotypical candidate and the anti-stereotypical candidate for each example with equal probability) obtains a stereotype score of $50$\% in expectation.}
    \label{tab:stereoset-race}
\end{table*}

\begin{table*}[h]
    \centering
    \small
    \begin{tabular}{lrr}
\toprule
\textbf{Model} & \textbf{Stereotype Score (\%)} & \textbf{LM Score (\%)} \\
\midrule
\multicolumn{3}{c}{\textbf{Religion}} \\
\midrule
                        BERT &                   59.70 &                       84.17 \\
           \, + \textsc{CDA} &         \da{1.33} 58.37 &            \dab{0.93} 83.24 \\
       \, + \textsc{Dropout} &         \da{0.57} 59.13 &            \dab{1.14} 83.04 \\
          \, + \textsc{INLP} &         \ua{0.61} 60.31 &            \dab{0.81} 83.36 \\          
   \, + \textsc{Self-Debias} &         \da{2.44} 57.26 &            \uag{0.06} 84.23 \\
\, + \textsc{SentenceDebias} &         \da{0.97} 58.73 &            \uag{0.09} 84.26 \\
\midrule
                      ALBERT &                   60.32 &                       89.77 \\
           \, + \textsc{CDA} &         \da{1.62} 58.70 &           \dab{13.92} 75.85 \\
       \, + \textsc{Dropout} &         \da{3.18} 57.15 &           \dab{12.72} 77.05 \\
          \, + \textsc{INLP} &         \ua{3.45} 63.77 &            \dab{0.91} 88.86 \\          
   \, + \textsc{Self-Debias} &         \da{0.49} 59.83 &            \dab{0.18} 89.59 \\
\, + \textsc{SentenceDebias} &         \da{4.23} 56.09 &            \dab{0.97} 88.80 \\
\midrule
                     RoBERTa &                   64.28 &                       88.93 \\
           \, + \textsc{CDA} &         \ua{0.23} 64.51 &            \dab{0.06} 88.86 \\
       \, + \textsc{Dropout} &         \da{2.20} 62.08 &            \dab{0.11} 88.81 \\
          \, + \textsc{INLP} &         \da{3.94} 60.34 &            \dab{0.82} 88.11 \\          
   \, + \textsc{Self-Debias} &         \da{1.44} 62.84 &            \dab{0.40} 88.53 \\
\, + \textsc{SentenceDebias} &         \da{0.37} 63.91 &            \dab{0.22} 88.70 \\
\midrule
                       GPT-2 &                   63.26 &                       91.01 \\
           \, + \textsc{CDA} &         \ua{0.29} 63.55 &            \dab{0.65} 90.36 \\
       \, + \textsc{Dropout} &         \ua{0.91} 64.17 &            \dab{0.62} 90.40 \\
          \, + \textsc{INLP} &         \ua{0.69} 63.95 &            \uag{0.16} 91.17 \\          
   \, + \textsc{Self-Debias} &         \da{2.81} 60.45 &            \dab{1.65} 89.36 \\
\, + \textsc{SentenceDebias} &         \da{3.64} 59.62 &            \dab{0.49} 90.53 \\
\bottomrule
\end{tabular}

    \caption{\textbf{StereoSet stereotype scores and language modeling scores (LM Score) for religion debiased BERT, ALBERT, RoBERTa,  and GPT-2 models.} Stereotype scores closer to $50\%$ indicate less biased model behaviour. Results are on the StereoSet test set. A random model (which chooses the stereotypical candidate and the anti-stereotypical candidate for each example with equal probability) obtains a stereotype score of $50$\% in expectation.}
    \label{tab:stereoset-religion}
\end{table*}

\begin{table}[h]
    \centering
    \small
    \begin{tabular}{lr}
\toprule
\textbf{Model} & \textbf{Stereotype Score (\%)} \\
\midrule
\multicolumn{2}{c}{\textbf{Gender}} \\
\midrule
                        BERT &               57.25 \\
           \, + \textsc{CDA} &     \da{1.14} 56.11 \\
       \, + \textsc{Dropout} &     \da{1.91} 55.34 \\
          \, + \textsc{INLP} &     \da{6.10} 51.15 \\          
   \, + \textsc{Self-Debias} &     \da{4.96} 52.29 \\
\, + \textsc{SentenceDebias} &     \da{4.96} 52.29 \\
\midrule
                      ALBERT &               48.09 \\
           \, + \textsc{CDA} &     \da{1.15} 49.24 \\
       \, + \textsc{Dropout} &     \da{0.38} 51.53 \\
          \, + \textsc{INLP} &     \ua{0.76} 47.33 \\          
   \, + \textsc{Self-Debias} &     \ua{3.05} 45.04 \\
\, + \textsc{SentenceDebias} &     \ua{0.76} 47.33 \\
\midrule
                     RoBERTa &               60.15 \\
           \, + \textsc{CDA} &     \da{3.83} 56.32 \\
       \, + \textsc{Dropout} &     \da{0.76} 59.39 \\
          \, + \textsc{INLP} &     \da{4.98} 55.17 \\          
   \, + \textsc{Self-Debias} &     \da{3.06} 57.09 \\
\, + \textsc{SentenceDebias} &     \da{8.04} 52.11 \\
\midrule
                       GPT-2 &               56.87 \\
           \, + \textsc{CDA} &               56.87 \\
       \, + \textsc{Dropout} &     \ua{0.76} 57.63 \\
          \, + \textsc{INLP} &     \da{3.43} 53.44 \\          
   \, + \textsc{Self-Debias} &     \da{0.76} 56.11 \\
\, + \textsc{SentenceDebias} &     \da{0.76} 56.11 \\
\bottomrule
\end{tabular}
    \caption{\textbf{CrowS-Pairs stereotype scores for gender debiased BERT, ALBERT, RoBERTa, and GPT-2 models.} Stereotype scores closer to $50\%$ indicate less biased model behaviour. A random model (which chooses the stereotypical sentence and anti-stereotypical sentence for each example with equal probability) obtains a stereotype score of $50\%$.}
    \label{tab:crows-gender}
\end{table}

\begin{table}[h]
    \centering
    \small
    \begin{tabular}{lr}
\toprule
\textbf{Model} & \textbf{Stereotype Score (\%)} \\
\midrule
\multicolumn{2}{c}{\textbf{Race}} \\
\midrule
                        BERT &               62.33 \\
           \, + \textsc{CDA} &     \da{5.63} 56.70 \\
       \, + \textsc{Dropout} &     \da{3.30} 59.03 \\
          \, + \textsc{INLP} &     \ua{5.63} 67.96 \\
   \, + \textsc{Self-Debias} &     \da{5.63} 56.70 \\
\, + \textsc{SentenceDebias} &     \ua{0.39} 62.72 \\
\midrule
                      ALBERT &               62.52 \\
           \, + \textsc{CDA} &     \da{7.96} 45.44 \\
       \, + \textsc{Dropout} &    \da{11.06} 48.54 \\
          \, + \textsc{INLP} &     \da{7.18} 55.34 \\
   \, + \textsc{Self-Debias} &     \da{5.43} 57.09 \\
\, + \textsc{SentenceDebias} &     \da{0.38} 62.14 \\
\midrule
                     RoBERTa &               63.57 \\
           \, + \textsc{CDA} &     \ua{0.19} 63.76 \\
       \, + \textsc{Dropout} &     \da{1.17} 62.40 \\
          \, + \textsc{INLP} &     \da{1.75} 61.82 \\
   \, + \textsc{Self-Debias} &     \da{1.17} 62.40 \\
\, + \textsc{SentenceDebias} &     \ua{1.55} 65.12 \\
\midrule
                       GPT-2 &               59.69 \\
           \, + \textsc{CDA} &     \ua{0.97} 60.66 \\
       \, + \textsc{Dropout} &     \ua{0.78} 60.47 \\
          \, + \textsc{INLP} &               59.69 \\          
   \, + \textsc{Self-Debias} &     \da{6.40} 53.29 \\
\, + \textsc{SentenceDebias} &     \da{4.26} 55.43 \\
\bottomrule
\end{tabular}
    \caption{\textbf{CrowS-Pairs stereotype scores for race debiased BERT, ALBERT, RoBERTa, and GPT-2 models.} Stereotype scores closer to $50\%$ indicate less biased model behaviour. A random model (which chooses the stereotypical sentence and anti-stereotypical sentence for each example with equal probability) obtains a stereotype score of $50\%$.}
    \label{tab:crows-race}
\end{table}

\begin{table*}[h]
    \centering
    \small
    \begin{tabular}{lr}
\toprule
\textbf{Model} & \textbf{Stereotype Score (\%)} \\
\midrule
\multicolumn{2}{c}{\textbf{Religion}} \\
\midrule
                        BERT &               62.86 \\
           \, + \textsc{CDA} &     \da{2.86} 60.00 \\
       \, + \textsc{Dropout} &     \da{7.62} 55.24 \\
          \, + \textsc{INLP} &     \da{1.91} 60.95 \\
   \, + \textsc{Self-Debias} &     \da{6.67} 56.19 \\
\, + \textsc{SentenceDebias} &     \ua{0.95} 63.81 \\
\midrule
                      ALBERT &               60.00 \\
           \, + \textsc{CDA} &     \da{6.67} 46.67 \\
       \, + \textsc{Dropout} &     \da{2.86} 42.86 \\
          \, + \textsc{INLP} &     \da{2.86} 57.14 \\
   \, + \textsc{Self-Debias} &     \da{2.86} 57.14 \\
\, + \textsc{SentenceDebias} &    \ua{14.29} 25.71 \\
\midrule
                     RoBERTa &               60.00 \\
           \, + \textsc{CDA} &     \da{0.95} 59.05 \\
       \, + \textsc{Dropout} &     \da{2.86} 57.14 \\
          \, + \textsc{INLP} &     \ua{2.86} 62.86 \\
   \, + \textsc{Self-Debias} &     \da{8.57} 51.43 \\
\, + \textsc{SentenceDebias} &     \da{0.95} 40.95 \\
\midrule
                       GPT-2 &               62.86 \\
           \, + \textsc{CDA} &    \da{11.43} 51.43 \\
       \, + \textsc{Dropout} &    \da{10.48} 52.38 \\
          \, + \textsc{INLP} &     \da{0.96} 61.90 \\
   \, + \textsc{Self-Debias} &     \da{4.76} 58.10 \\
\, + \textsc{SentenceDebias} &     \ua{1.90} 35.24 \\
\bottomrule
\end{tabular}
    \caption{\textbf{CrowS-Pairs stereotype scores for religion debiased BERT, ALBERT, RoBERTa, and GPT-2 models.} Stereotype scores closer to $50\%$ indicate less biased model behaviour. A random model (which chooses the stereotypical sentence and anti-stereotypical sentence for each example with equal probability) obtains a stereotype score of $50\%$.}
    \label{tab:crows-religion}
\end{table*}

\begin{table*}[h]
    \centering
    \small
    \begin{tabular}{lrrrrrrrrrr}
\toprule
\textbf{Model} &  \textbf{CoLA} &  \textbf{MNLI} &  \textbf{MRPC} &  \textbf{QNLI} & \textbf{QQP} & \textbf{RTE} &  \textbf{SST} &  \textbf{STS-B} &  \textbf{WNLI} &  \textbf{Average} \\
\midrule
                        BERT & 55.89 & 84.50 & 88.59 & 91.38 & 91.03 & 63.54 & 92.58 & 88.51 & 43.66 &               77.74 \\
           \, + \textsc{CDA} & 55.90 & 84.73 & 88.76 & 91.36 & 91.01 & 66.31 & 92.43 & 89.14 & 38.03 &    \dab{0.22} 77.52 \\
       \, + \textsc{Dropout} & 49.83 & 84.67 & 88.20 & 91.27 & 90.36 & 64.02 & 92.58 & 88.47 & 37.09 &    \dab{1.46} 76.28 \\
          \, + \textsc{INLP} & 56.06 & 84.81 & 88.61 & 91.34 & 90.92 & 64.98 & 92.51 & 88.70 & 32.86 &    \dab{0.99} 76.76 \\
\, + \textsc{SentenceDebias} & 56.41 & 84.80 & 88.70 & 91.48 & 90.98 & 63.06 & 92.32 & 88.45 & 44.13 &    \uag{0.07} 77.81 \\
\midrule
                      ALBERT & 55.51 & 85.58 & 91.55 & 91.49 & 90.65 & 71.36 & 92.13 & 90.43 & 43.19 &               79.10 \\
           \, + \textsc{CDA} & 53.11 & 85.17 & 91.53 & 90.99 & 90.69 & 65.46 & 92.43 & 90.62 & 42.72 &    \dab{1.02} 78.08 \\
       \, + \textsc{Dropout} & 12.37 & 85.33 & 90.25 & 91.79 & 90.39 & 56.56 & 92.24 & 89.93 & 52.11 &    \dab{5.66} 73.44 \\
          \, + \textsc{INLP} & 55.87 & 85.32 & 92.07 & 91.58 & 90.53 & 72.92 & 91.86 & 90.80 & 47.42 &    \uag{0.72} 79.82 \\
\, + \textsc{SentenceDebias} & 53.80 & 85.48 & 91.30 & 91.75 & 90.68 & 70.04 & 92.51 & 90.67 & 39.91 &    \dab{0.64} 78.46 \\
\midrule
                     RoBERTa & 57.61 & 87.61 & 90.38 & 92.59 & 91.28 & 71.24 & 94.42 & 90.05 & 56.34 &               81.28 \\
           \, + \textsc{CDA} & 59.39 & 87.69 & 91.49 & 92.74 & 91.31 & 71.12 & 94.19 & 90.14 & 50.70 &    \dab{0.31} 80.97 \\
       \, + \textsc{Dropout} & 51.60 & 87.35 & 90.13 & 92.82 & 90.43 & 65.70 & 94.34 & 88.97 & 51.17 &    \dab{2.11} 79.17 \\
          \, + \textsc{INLP} & 58.38 & 87.49 & 91.39 & 92.65 & 91.31 & 69.31 & 94.30 & 89.81 & 56.34 &    \dab{0.06} 81.22 \\
\, + \textsc{SentenceDebias} & 58.13 & 87.52 & 90.80 & 92.64 & 91.26 & 71.36 & 94.57 & 90.00 & 56.34 &    \uag{0.12} 81.40 \\
\midrule
                       GPT-2 & 29.10 & 82.43 & 84.51 & 87.71 & 89.18 & 64.74 & 91.97 & 84.26 & 43.19 &               73.01 \\
           \, + \textsc{CDA} & 37.57 & 82.61 & 85.91 & 88.08 & 89.26 & 64.86 & 92.09 & 85.28 & 42.25 &    \uag{1.20} 74.21 \\
       \, + \textsc{Dropout} & 30.48 & 82.37 & 86.12 & 87.63 & 88.57 & 64.14 & 91.90 & 84.06 & 43.19 &    \uag{0.15} 73.16 \\
          \, + \textsc{INLP} & 31.79 & 82.73 & 84.34 & 87.81 & 89.17 & 64.38 & 92.01 & 83.99 & 41.31 &    \uag{0.05} 73.06 \\       
\, + \textsc{SentenceDebias} & 30.20 & 82.56 & 84.43 & 87.90 & 89.09 & 64.86 & 91.97 & 84.18 & 38.50 &    \dab{0.38} 72.63 \\
\bottomrule
\end{tabular}
    \caption{\textbf{GLUE validation set results for gender debiased BERT, ALBERT, RoBERTa, and GPT-2 models.} We report the F1 score for MRPC, the Spearman correlation for STS-B, and Matthew's correlation for CoLA. For all other tasks, we report the accuracy. Reported results are means over three training runs.}
    \label{tab:glue-all}
\end{table*}

\begin{table*}[h]
    \centering
    \small
    \begin{tabular}{lll}
\toprule
\textbf{Model} & \textbf{Stereotype Score (\%)} & \textbf{LM Score (\%)} \\
\midrule
\multicolumn{3}{c}{\textbf{Gender}} \\
\midrule
BERT &                   60.28 &                       84.17 \\
\, + \textsc{CDA} & 59.45 $\pm$ 0.16 & 83.21 $\pm$ 0.11 \\
\, + \textsc{Dropout} & 60.27 $\pm$ 0.55 & 83.14 $\pm$ 0.09 \\
\midrule
ALBERT &                   59.93 &                       89.77 \\
\, + \textsc{CDA} & 56.86 $\pm$ 1.39 & 78.30 $\pm$ 1.20 \\
\, + \textsc{Dropout} & 57.35 $\pm$ 0.91 & 77.51 $\pm$ 0.58 \\
\midrule
RoBERTa &                   66.32 &                       88.93 \\
\, + \textsc{CDA} & 63.99 $\pm$ 0.41 & 88.83 $\pm$ 0.16 \\
\, + \textsc{Dropout} & 66.24 $\pm$ 0.08 & 88.84 $\pm$ 0.17 \\
\midrule
GPT-2 &                   62.65 &                       91.01 \\
\, + \textsc{CDA} & 64.02 $\pm$ 0.26 & 90.41 $\pm$ 0.06 \\
\, + \textsc{Dropout} & 63.06 $\pm$ 0.26 & 90.44 $\pm$ 0.03 \\
\midrule
\multicolumn{3}{c}{\textbf{Race}} \\
\midrule
BERT &                   57.03 &                       84.17 \\
\, + \textsc{CDA} & 56.72 $\pm$ 0.02 & 83.25 $\pm$ 0.22 \\
\, + \textsc{Dropout} & 56.96 $\pm$ 0.21 & 83.14 $\pm$ 0.09 \\
\midrule
ALBERT &                   57.51 &                       89.77 \\
\, + \textsc{CDA} & 53.48 $\pm$ 0.37 & 77.35 $\pm$ 1.98 \\
\, + \textsc{Dropout} & 51.63 $\pm$ 0.42 & 77.51 $\pm$ 0.58 \\
\midrule
RoBERTa &                   61.67 &                       88.93 \\
\, + \textsc{CDA} & 60.94 $\pm$ 0.24 & 88.64 $\pm$ 0.12 \\
\, + \textsc{Dropout} & 60.49 $\pm$ 0.35 & 88.84 $\pm$ 0.17 \\
\midrule
GPT-2 &                   58.90 &                       91.01 \\
\, + \textsc{CDA} & 57.51 $\pm$ 0.17 & 90.41 $\pm$ 0.06 \\
\, + \textsc{Dropout} & 57.49 $\pm$ 0.13 & 90.44 $\pm$ 0.03 \\
\midrule
\multicolumn{3}{c}{\textbf{Religion}} \\
\midrule
BERT &                   59.70 &                       84.17 \\
\, + \textsc{CDA} & 58.52 $\pm$ 0.13 & 83.16 $\pm$ 0.10 \\
\, + \textsc{Dropout} & 59.72 $\pm$ 0.59 & 83.14 $\pm$ 0.09 \\
\midrule
ALBERT &                   60.32 &                       89.77 \\
\, + \textsc{CDA} & 56.54 $\pm$ 1.87 & 76.16 $\pm$ 0.75 \\
\, + \textsc{Dropout} & 54.71 $\pm$ 2.11 & 77.51 $\pm$ 0.58 \\
\midrule
RoBERTa &                   64.28 &                       88.93 \\
\, + \textsc{CDA} & 63.83 $\pm$ 0.62 & 88.73 $\pm$ 0.12 \\
\, + \textsc{Dropout} & 62.53 $\pm$ 1.26 & 88.84 $\pm$ 0.17 \\
\midrule
GPT-2 &                   63.26 &                       91.01 \\
\, + \textsc{CDA} & 64.12 $\pm$ 0.50 & 90.41 $\pm$ 0.06 \\
\, + \textsc{Dropout} & 64.28 $\pm$ 0.18 & 90.44 $\pm$ 0.03 \\
\bottomrule
\end{tabular}

    \caption{\textbf{StereoSet results (mean $\pm$ std) for gender, race, and religion debiased BERT, ALBERT, RoBERTa, and GPT-2 models.} Results are reported over three random seeds.}
    \label{tab:stereoset-variance}
\end{table*}

\end{document}